# Weight-Space Programming for Targeted Repair of Robot Foundation Models

Kyle Coelho, Eugen Solowjow
Siemens Research
kyle.coelho@siemens.com

## Introduction

Weight-space programming (WSP) asks whether a language-model agent can inspect and directly write a neural network's parameters, instead of adapting it through gradient descent. MIT's initial WSP report [1] brought this idea into view, showing that frontier LLMs such as GPT Astra can directly read, edit, and write the weights of neural networks to achieve a desired behavior and end goal. This was showcased on vision tasks and math problem solving, among others. Within two weeks of the authors' post, the results attracted considerable attention due to the potentially profound implications of WSP. We report the application of WSP to robot foundation models (RFMs) and show how failure modes of an RFM can be mitigated without additional data collection or training.

We test whether failures of RFM policies can be corrected through direct weight editing. Our hypothesis is that many failures are not random: the policy repeatedly lands in the same recognizable stuck state, such as hovering above a target or being unable to reach a target pose. If that situation is recognizable in the model's internal activations, a small edit can use that signal to trigger a correction while leaving the other model parameters unchanged. Prior work has directly edited robot policies in weight space. Stranghöner et al. wrote state-dependent controllers into unused units of behavior-cloning policies [2]. Other work has edited Pi0.5 [3] to change action execution speed [4] or repair instruction grounding [5, 6]. We investigate recurring physical execution failures in a Pi0.5 policy. Our edits write task-specific corrective movements into the action expert that activate in the diagnosed failure states. Paired held-out evaluations, including random and sign-reversed controls, test whether the edits improve performance through the intended correction. We also evaluate an edit on a policy already fine-tuned for its task.

## Simulation Environment, Policies, and Editing Agent

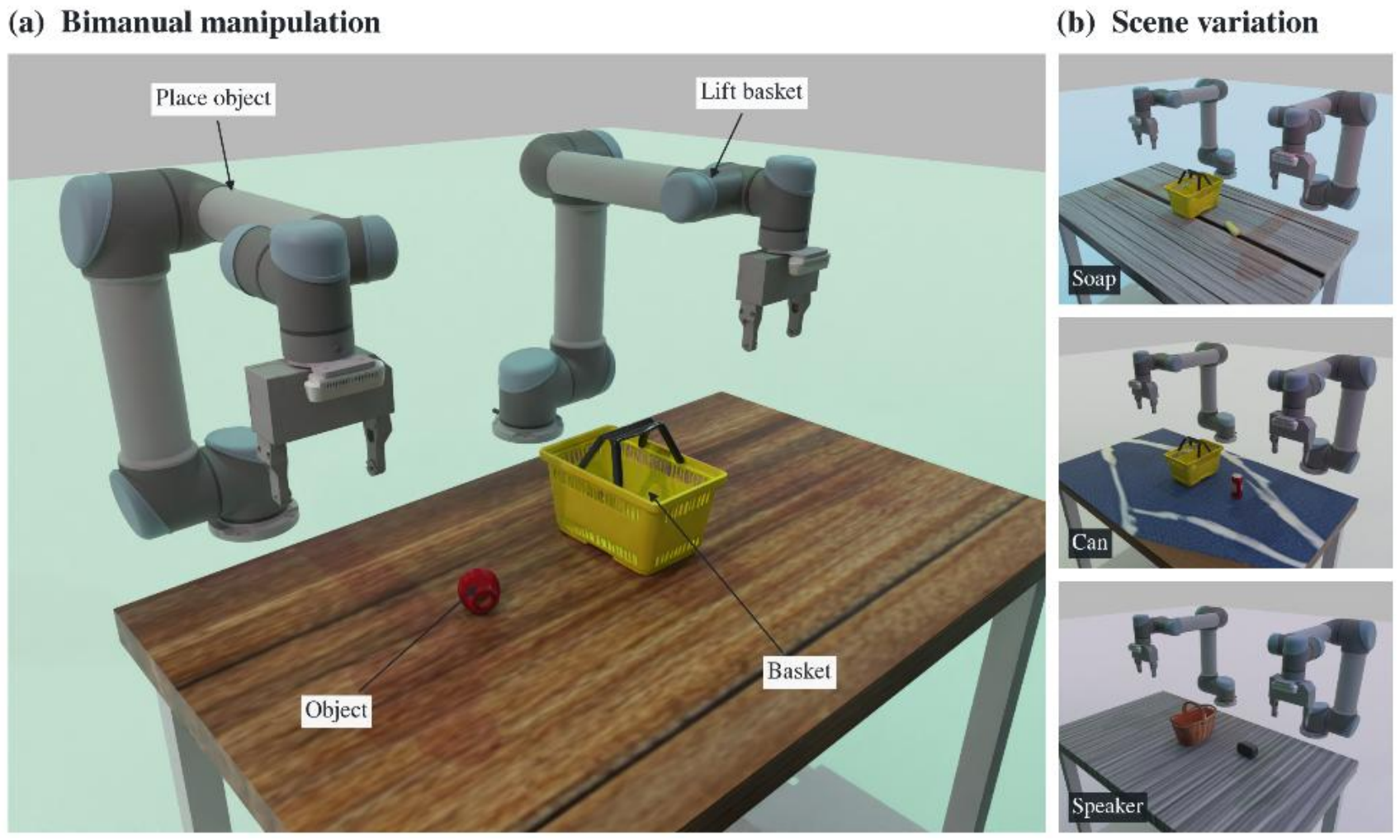

*Fig 1. (A) RoboTwin bimanual UR5 setup with WSG grippers for the place_mixed_objects_basket task. (B) Scenes vary in object, basket, texture, lighting, and background.*

We use the RoboTwin 2.0 simulator [7] where, in a closed loop, the policy reads camera images and arm state, plans its next 50 movements, executes them, and repeats. An agent (Claude Opus 5.5 with tool access) starts and stops the policy server, runs the policy on tasks, inspects the actions produced and the neurons active at each step, edits the weights, and tests each edit in simulation. Because the model is too large for the agent's context, the agent examines only targeted parts of the vision-language-action (VLA) model. No new demonstrations were collected and no training (supervised fine-tuning, reinforcement learning, or gradient updates) was used. Original and edited policies were run on identical seeded scenes, and each edit was judged on new scenes against a pass mark set in advance.

## Target Failure Modes and Edit Procedure

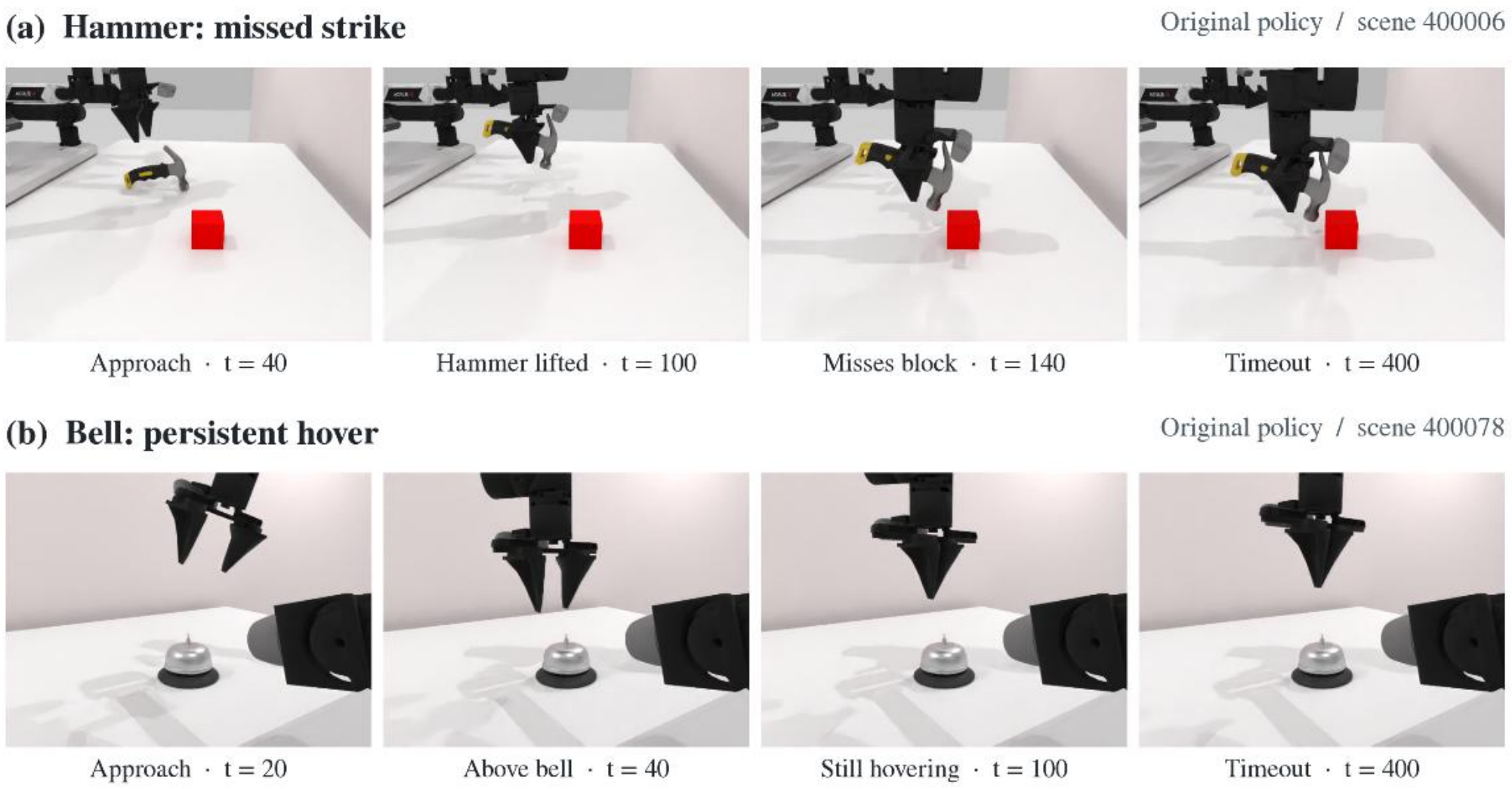

*Fig 2. Each row shows the unedited policy at four steps from one held-out rollout, progressing from approach to failure. (A) The hammer strike misses the block and the robot stalls. (B) In the bell task, the gripper remains above the bell without making contact.*

We ran the unedited RoboTwin Pi0.5 policy on a range of tasks and selected four with consistently recurring failures:

- Hammer to Block (beat_block_hammer): The hammer swivels slightly in the gripper, the strike lands 2 - 4.5 cm off the block, and the robot freezes instead of correcting. This occurred in 43% of runs.
- Click Bell (click_bell): In about 1 in 10 runs, the closed gripper stops a few centimeters above the bell and hovers without descending.
- Can into Pot (move_can_pot): In about a third of runs, the gripper closes too high on the can, which twists out of the grasp when lifted.
- Turn Switch (turn_switch): In almost half of runs, the gripper presses 2-4.6 cm below the switch and jams against its box.

We also ran a checkpoint specifically fine-tuned (full-weight fine-tuning for 30k steps) on the following task to test whether our approach improves a model that has already been specialized for a task:

- Objects into Basket (place_mixed_objects_basket): The gripper closes beside the object instead of around it and lifts nothing. This occurred in about 1 in 6 runs (84 of 505), accounting for about half of all failures (84 of 163) (Appendix Fig 5).

Each fix followed the same procedure. The agent identified each failure from the simulator's exact measurements, then recorded the internal activity of Pi0.5's action expert (which translates the model's visual understanding of the scene into movements) to find a pattern that marks the stuck state. In the base model, a survey of all 73,728 neurons in the module's action expert found only six that were essentially never active on the tasks checked, all in its final layer. The fine-tuned basket checkpoint had two similarly near-silent neurons in its final layer, and the basket fix used those. Each fix rewrites connections to two near-silent neurons, one for each arm, so the added correction activates in a situation associated with the diagnosed failure. For example, the hammer edit nudges the hammer toward the block after a miss, the bell edit lowers the gripper toward the bell, and the basket edit shifts the gripper during its final approach to the object.

The fixes change only about 6,000 of the model's 3.35 billion parameters and leave its vision-language parts untouched. Each edit was tried on an initial set of scenes, revised if needed, confirmed on separate validation scenes, then locked and tested on new scenes (200 for beat_block_hammer, 300 for click_bell, 200 for move_can_pot, 505 for place_mixed_objects_basket; turn_switch did not reach this stage). The hammer task, for instance, took four iterations. The first three either barely changed the robot's actions or caused new misses; the fourth raised success on 200 new scenes from 56.5% to 70.5%.

## Results

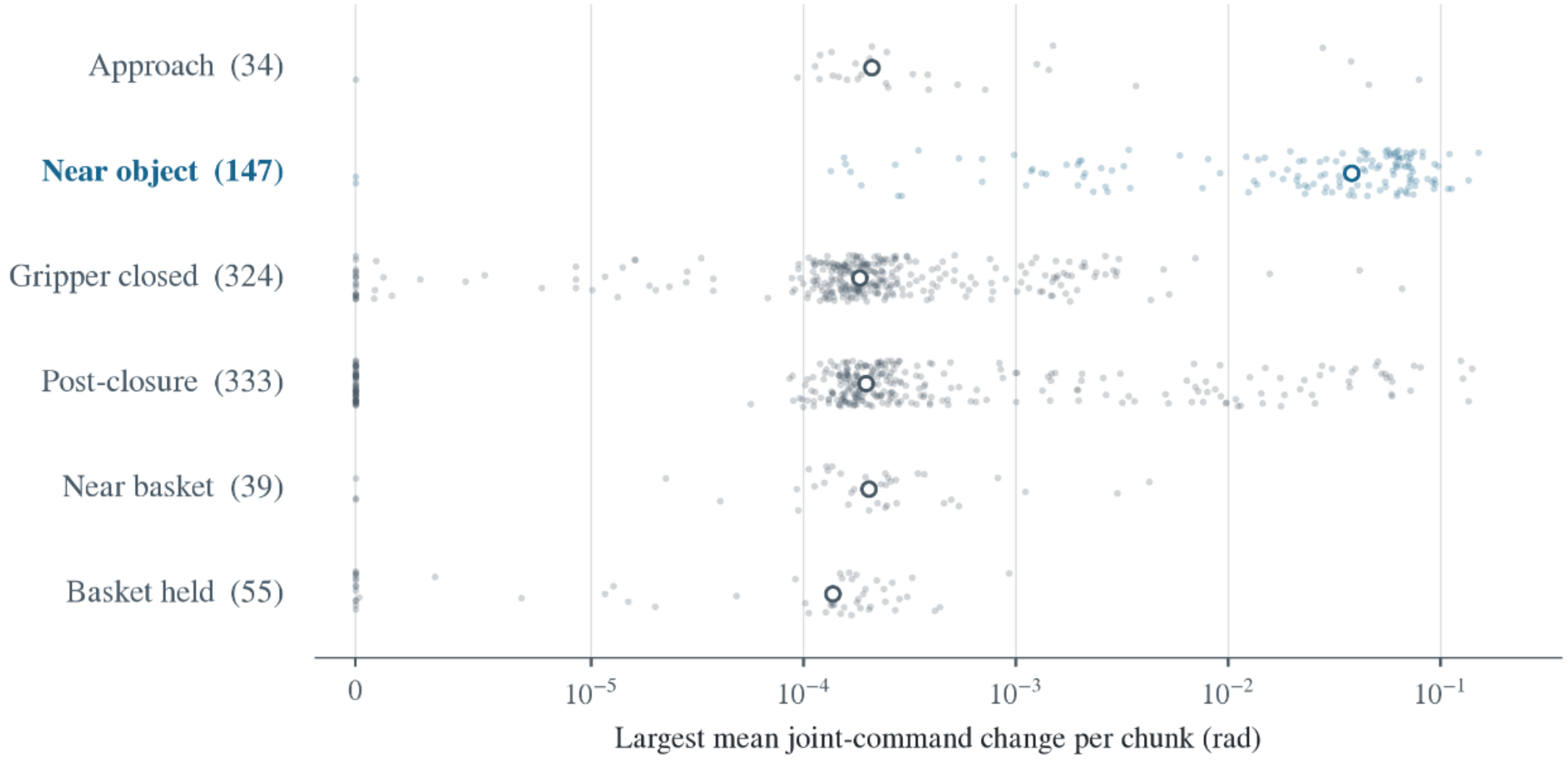

*Fig 3. Impact of edit changes on joint commands at each stage of the place_mixed_objects_basket task given the same recorded inputs from 30 episodes. Each dot is one action chunk and shows the biggest change across the 12 arm-joint commands, averaged over the chunk. Open circles are phase median values. The median change near an object is about 180–280× larger than in any other phase.*

Three of the five fixes clearly worked on unseen scenes. On the hammer task, success rose from 56.5% to 70.5% across 200 new scenes: the edit fixed 41 scenes but broke 13. Recovery after a missed strike rose from 22% to 48% of misses, mostly from the right arm. On the bell task, success rose from 87.3% to 96.7% across 300 new scenes: 29 scenes were fixed and 1 was broken. Hovering failures fell from 11 to 2, while median time to success remained similar (Appendix Fig 4).

On the already task-fine-tuned basket model, success rose from 67.7% to 84.4% across 505 new scenes: the edit fixed 121 scenes and broke 37. Missed grasps, where the gripper touched the object but lifted nothing, fell from 84 to 12. Scenes solved by both versions were completed just as quickly. Direct weight editing can therefore further improve an already task fine-tuned policy (Appendix Fig 4).

The edited behavior, not a lucky disturbance of the model, explains these results. Random edits of the same size made no consistent difference. Reversing the bell edit, so that it lifts instead of lowers, created the very failure it was meant to fix: on the first 100 test scenes, hovering failures rose from 4 to 19, while successes fell from 91 to 64. Reversing the hammer edit moved the hammer head the opposite way and likewise removed the gain. Reversing the basket edit made the gripper close even further short of the object, about 5 cm instead of 3 cm, and cut success from 70 to 34 out of 100.

For the other two tasks, the move_can_pot edit lowered the robot's approach as designed, but not enough: success went from 38% to 41% on 200 new scenes, a difference that is not statistically significant. The turn_switch edits likewise showed no significant improvement to advance to held-out testing.

Although the basket edits targeted only missed grasps, failures in which the object was lifted but ended up back on the table instead of in the basket fell from 38 to 17. Failures at the basket stage rose slightly, from 12 to 19, though many more objects now reach that stage. Two object types (cups and playing cards) did slightly worse, though not significantly.

## Conclusion and Outlook

We applied direct weight editing of robot foundation models using frontier LLMs. The reported experiments showed that small, targeted weight edits can improve some recurring failures of large-scale robot foundation models in closed loop, without new demonstrations or training. The working edits changed about 6,000 parameters, roughly 0.0002% of the model, and added task-specific corrective movements. The same kind of edit also improved a policy that had already been fully fine-tuned on its task. The approach worked best when the failure situation was recognizable and a simple correction could address it; it did not solve every failure.

Weight-space programming of robot foundation models is at a very early stage (as is WSP in general), and its practical scope is still largely unexplored. Yet early results suggest that manipulating models directly in weight space could provide a fundamentally different mechanism for adapting and understanding robotic capabilities. Establishing when such operations are reliable and predictable could therefore shape how future robot foundation models are developed and programmed. By uncovering structure in weight space and relating it to specific behaviors, WSP could make the inner workings of large robotic models more transparent.

## Appendix

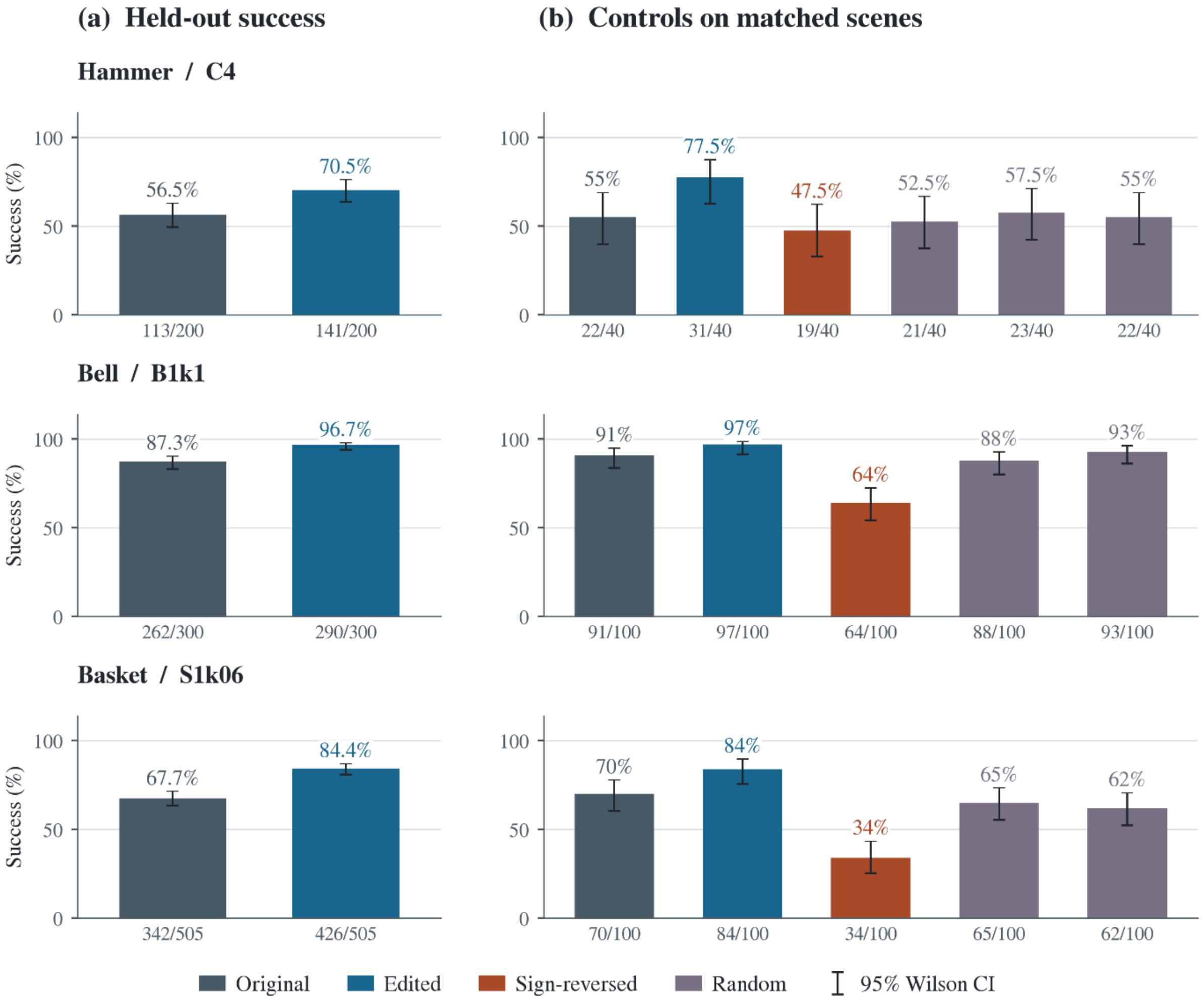

*Fig 4. Each row represents a single task. Column A shows the direct success rate improvement; Column B shows control runs with edited, reversed, and random weights. C4, B1k1, and S1k06 are groups of model parameters edited in the action expert.*

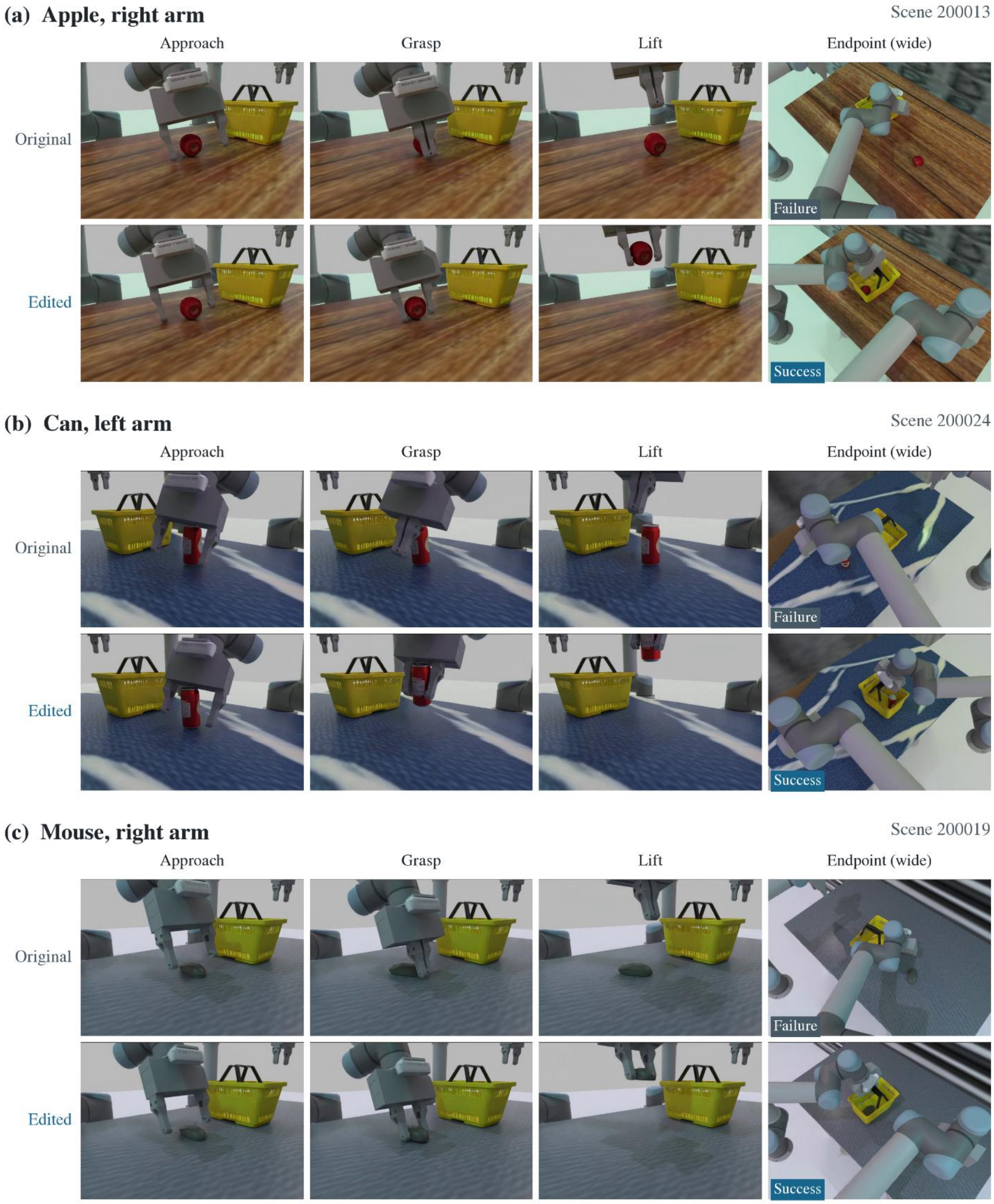

*Fig 5. Each row shows a comparison between the original supervised fine-tuned (SFT) and edited model on the place_mixed_objects_basket task. The last column in each row represents each policy's final timestamp.*